\documentclass{article}
\PassOptionsToPackage{numbers, compress}{natbib}
\usepackage[final,sglblindworkshop]{neurips_2026}

\usepackage[utf8]{inputenc}
\usepackage[T1]{fontenc}
\usepackage[hidelinks]{hyperref}
\usepackage{url}
\usepackage{booktabs}
\usepackage{amsmath,amssymb,amsfonts}
\usepackage{nicefrac}
\usepackage{microtype}
\usepackage{xcolor}
\usepackage{xspace}
\usepackage{graphicx}
\usepackage{multirow}
\usepackage{enumitem}
\usepackage{mathtools}
\usepackage{cleveref}
\usepackage{tikz}
\usetikzlibrary{arrows.meta,positioning,fit}

\newcommand{\RippleCP}{\textsc{RippleCP}\xspace}
\newcommand{\QRipple}{\textsc{Q-Ripple}\xspace}

\newcommand{\placeholderfigure}[2]{%
  \fbox{\parbox[c][#2][c]{0.95\linewidth}{\centering\small\textbf{FIGURE PLACEHOLDER}\\[1.5mm]#1}}
}
\newcommand{\maybefigure}[4]{%
  \IfFileExists{figures/#1}{%
    \includegraphics[width=#2]{figures/#1}%
  }{%
    \placeholderfigure{Add \nolinkurl{figures/#1}.\\#4}{#3}%
  }%
}

\newcommand{\NumTasks}{12 tasks\xspace}
\newcommand{\NumTasksN}{12}

\newcommand{\NumTrajectories}{12 nominal trajectories\xspace}
\newcommand{\NumCandidates}{24 selected checkpoint cells\xspace}
\newcommand{\NumCandidatesN}{24}
\newcommand{\NumNominalCandidates}{27}
\newcommand{\NumCounterfactuals}{53 matched CP/SKIP pairs, comprising 106 recovery branches\xspace}
\newcommand{\NumPairs}{53}
\newcommand{\NumBranches}{106}

\newcommand{\FailureWindow}{one early and one late checkpoint boundary per task, followed by the next logical failure step\xspace}
\newcommand{\PrimaryAgentModel}{Codex CLI 0.147.0-alpha.6.5, GPT-5.4, xhigh reasoning\xspace}
\newcommand{\CheckpointBackend}{archive-backed logical checkpoints containing workspace files and preserved agent-visible state\xspace}
\newcommand{\CheckpointBackendShort}{archive-backed logical workspace and agent-state checkpoints\xspace}

\newcommand{\NominalSuccess}{10/12\xspace}

\newcommand{\StabilizedCells}{5}

\newcommand{\VerCP}{100.7}
\newcommand{\VerSKIP}{150.2}
\newcommand{\VerAdvantage}{49.4}
\newcommand{\VerAdvantageCI}{34.1--65.5}

\newcommand{\VerTasks}{10}

\newcommand{\VerTokenAdvantage}{203{,}171}
\newcommand{\VerToolCallAdvantage}{7.5}

\newcommand{\AllCP}{105.4}
\newcommand{\AllSKIP}{166.7}
\newcommand{\AllAdvantage}{61.3}
\newcommand{\AllAdvantageCI}{40.4--88.5}

\newcommand{\PositiveCells}{17}

\newcommand{\DjangoAdvantageVer}{52.6}
\newcommand{\SklearnAdvantageVer}{43.5}
\newcommand{\SympyAdvantageVer}{51.2}

\newcommand{\DjangoCP}{105.0}
\newcommand{\DjangoSKIP}{157.6}
\newcommand{\DjangoAdvantage}{52.6}
\newcommand{\SklearnCP}{116.8}
\newcommand{\SklearnSKIP}{165.4}
\newcommand{\SklearnAdvantage}{48.5}
\newcommand{\SympyCP}{94.3}
\newcommand{\SympySKIP}{177.1}
\newcommand{\SympyAdvantage}{82.7}

\newcommand{\CheckpointLatency}{4.38\,s mean (median 4.09\,s; range 2.50--6.68\,s)\xspace}
\newcommand{\CheckpointLatencyMean}{4.38}
\newcommand{\RestoreLatency}{9.69\,s mean (median 6.76\,s; p95 20.8\,s)\xspace}

\newcommand{\CheckpointBytes}{mean 13.94\,MB; median 10.44\,MB; range 7.25--23.35\,MB\xspace}

\newcommand{\CheckpointOverheadMean}{4.41}
\newcommand{\AgentTimeoutCap}{1800}
\newcommand{\LongestBranch}{463}

\newcommand{\TotalTokens}{28{,}125{,}711}
\newcommand{\AgentHours}{3.38}
\newcommand{\BranchHours}{3.85}

\newcommand{\BootstrapReps}{10{,}000}
\newcommand{\BootstrapSeed}{20260818}
\newcommand{\ConfidenceInterval}{95\% task-bootstrap CI\xspace}

\newcommand{\FirstCPCP}{100.8}
\newcommand{\FirstCPSKIP}{200.8}
\newcommand{\FirstCPAdvantage}{100.0}
\newcommand{\FirstCPAdvantageCI}{72.4--131.1}
\newcommand{\AddCPCP}{100.6}
\newcommand{\AddCPSKIP}{99.5}
\newcommand{\AddCPAdvantage}{$-1.1$}
\newcommand{\AddCPAdvantageCI}{$-9.5$--$8.4$}

\newcommand{\RestorePristine}{226.5}
\newcommand{\RestoreOne}{108.6}
\newcommand{\RestoreTwo}{103.3}
\newcommand{\RestoreThree}{95.6}

\newcommand{\BudgetOneTasks}{7}
\newcommand{\BudgetOneAllTasks}{9}
\newcommand{\BudgetOneFirst}{105.9}
\newcommand{\BudgetOneSecond}{155.9}

\newcommand{\TrajStepsMean}{3.25}
\newcommand{\TrajStepsRange}{3--4}
\newcommand{\CandidateSteps}{1--3}
\newcommand{\FailureSteps}{2--4}

\newcommand{\BudgetOneFirstAll}{112.1}
\newcommand{\BudgetOneSecondAll}{175.6}

\newcommand{\MidTasks}{3}
\newcommand{\MidBoundaryTwo}{86.2}
\newcommand{\MidBoundaryThree}{94.7}
\newcommand{\MidGap}{8.6}

\newcommand{\LaterWinsLateFailure}{5}

\newcommand{\FirstTokenSaving}{388{,}812}

\newcommand{\StepOneShare}{73.9\%}
\newcommand{\StepTwoShare}{19.5\%}
\newcommand{\StepThreeShare}{6.0\%}
\newcommand{\StepFourShare}{0.6\%}
\newcommand{\StepOneShareMean}{72.8\%}
\newcommand{\StepOneMean}{172.1}
\newcommand{\StepTwoMean}{45.5}
\newcommand{\StepThreeMean}{13.9}
\newcommand{\StepFourMean}{5.4}
\newcommand{\CumTwo}{93.4\%}
\newcommand{\CumThree}{99.4\%}

\newcommand{\FirstProtected}{156.5}
\newcommand{\FirstConversion}{0.64}
\newcommand{\AddProtected}{41.8}
\newcommand{\AddConversion}{$-0.03$}

\newcommand{\RecutItems}{350}
\newcommand{\RecutPerTask}{29.2}
\newcommand{\RecutTurnOne}{24.8}
\newcommand{\RecutTurnOneRange}{12--41}
\newcommand{\RecutCommands}{276}
\newcommand{\RecutFileChanges}{34}

\newcommand{\ProxyTool}{$+0.85$}
\newcommand{\ProxyToolP}{0.003}
\newcommand{\ProxyFiles}{$+0.71$}
\newcommand{\ProxyFilesP}{0.025}
\newcommand{\ProxyBytes}{$+0.70$}
\newcommand{\ProxyBytesP}{0.030}
\newcommand{\ProxyFacts}{$+0.43$}
\newcommand{\ProxyFactsP}{0.210}
\newcommand{\ProxyImportance}{$-0.58$}
\newcommand{\ProxyImportanceP}{0.082}
\newcommand{\ProxyToolDrop}{$+0.81$}
\newcommand{\ProxyBytesDrop}{$+0.83$}

\newcommand{\NoiseAll}{24.1}
\newcommand{\NoiseFirst}{33.1}
\newcommand{\NoiseFirstEffect}{121.9}
\newcommand{\NoiseFirstRatio}{0.27}
\newcommand{\NoiseAdd}{15.0}
\newcommand{\NoiseAddEffect}{14.4}
\newcommand{\NoiseAddRatio}{1.04}
\newcommand{\CostPerLabel}{8.4}

\newcommand{\ArchiveOverall}{16.1}
\newcommand{\ArchiveDjango}{1\%}
\newcommand{\ArchiveSklearn}{12\%}
\newcommand{\ArchiveSympy}{39\%}

\title{RippleCP: Measuring Counterfactual Checkpoint Advantage in Coding Agents}

\author{%
  \textbf{Mayur Akewar} \quad \textbf{Ravi Ranjan} \quad
  \\
  \texttt{makew001@fiu.edu} \quad \texttt{rkuma031@fiu.edu}
  \\
    \textit{Florida International University, Miami, USA} \\
}

\begin{document}
\maketitle

\begin{abstract}
Agent checkpoint systems decide what state is recovery-relevant, how to snapshot it, and whether rollback is admissible. None decides which of the safe boundaries they expose are worth materializing. We formulate this as \emph{counterfactual checkpoint advantage}, the reduction in future recovery cost obtained by checkpointing a candidate rather than skipping it, and measure it by driving a \textsc{CP} branch and a \textsc{Skip} branch to the same logical failure and recovering both under matched model, tool, verifier, and stopping conditions. On a frozen pilot of \NumTasksN{} SWE-bench Verified tasks and \NumBranches{} real recovery branches, checkpointing saves \VerAdvantage\,s per task, and that figure resolves into two regimes two orders of magnitude apart. The first checkpoint returns \FirstCPAdvantage\,s on \FirstProtected\,s of protected work, a conversion of \FirstConversion; a second one step later returns \AddCPAdvantage\,s on \AddProtected\,s, a conversion of \AddConversion. Recovery is a re-derivation rather than a replay, so preserved work is a poor guide to saved work, and the classical elapsed-work rule misprices the second checkpoint by its full nominal cost. We identify where placement can pay, and set the bar a placement policy must clear.
\end{abstract}

\section{Introduction}
Long-horizon agents increasingly behave like stateful systems: they alternate model calls with shell commands, file edits, tests, external tools, retries, and pauses. A late crash can force substantial re-execution, and a growing set of runtimes makes checkpoint and restore for such agents practical. \textsc{Crab} detects recovery-relevant OS effects~\citep{wu2026crab}, \textsc{DeltaBox} targets low-latency sandbox rollback~\citep{dong2026deltabox}, and \textsc{DART} reasons about semantically admissible recovery points~\citep{yang2026dart}. Each of them answers where a checkpoint \emph{may} be taken. None answers where one \emph{should} be: among the safe boundaries a runtime exposes, which are worth paying to make durable?

Classical placement answers that by trading checkpoint overhead against expected recomputation~\citep{young1974checkpoint}, which presumes recomputation cost is known. For agents it is not. Downloading a gigabyte may be cheap to repeat, while rediscovering a one-line dependency incompatibility can determine every subsequent edit, build, and test. We expect checkpoint-worthy agent state to be \emph{sparse in bytes but dense in downstream influence} (Appendix~\ref{app:intuition}), which would make the usual proxies of elapsed work and bytes changed unreliable. Whether it does is an empirical question, and Section~\ref{sec:pilot} answers it directly. Either way the quantity is one to measure rather than assume.

\RippleCP{} makes it measurable. Rather than labelling a state ``important,'' we run two matched futures from the same candidate: checkpoint now, or skip now. Both are driven by the real agent to the same logical failure and recovered under the same model, tool, verifier, and stopping conditions, so their cost difference, the \emph{checkpoint advantage}, comes from executed system behaviour rather than from an LLM judge or a human annotation. This paper reports that measurement on a frozen pilot; learning a policy from it is future work.

The result reshapes the placement question: the aggregate advantage is large, but it decomposes almost entirely into the value of \emph{having a checkpoint at all} rather than of \emph{choosing where to put it}. We contribute the advantage formulation, a paired-recovery harness producing labels from executed agent branches, and the pilot that separates those regimes and prices each.

\section{Checkpoint advantage formulation}
Let an agent produce states $s_0,\ldots,s_T$. At safe candidate $s_t$, the placement action is $a_t\in\{\mathrm{CP},\mathrm{SKIP}\}$. Let $c^{-}$ denote the latest materialized checkpoint before $t$, and let $f\sim\mathcal{F}$ denote a future failure. We define a cost-to-go under action $a$,
\begin{equation}
J(s_t,a)=\mathbb{E}_{f\sim\mathcal{F}}\!\left[C_{\mathrm{place}}(a)+L\big(\mathrm{Recover}(s_t,a,f)\big)\right],
\end{equation}
where $C_{\mathrm{place}}(\mathrm{SKIP})=0$, $C_{\mathrm{place}}(\mathrm{CP})=C_{\mathrm{ckpt}}(s_t)$, and $L$ is future recovery/re-execution loss. For connection to value-based decision making, define $Q(s_t,a)=-J(s_t,a)$. The \textbf{checkpoint advantage} is
\begin{align}
A_{\mathrm{CP}}(s_t)
&=Q(s_t,\mathrm{CP})-Q(s_t,\mathrm{SKIP}) \\
&=J(s_t,\mathrm{SKIP})-J(s_t,\mathrm{CP}).
\label{eq:advantage}
\end{align}
Thus $A_{\mathrm{CP}}(s_t)>0$ means checkpointing saves more future cost than it adds. Our primary loss is wall-clock redo time; repeated tokens, tool calls, and task success remain separate metrics rather than hidden scalar weights.

The formulation is independent of the snapshot mechanism: \RippleCP assumes candidates already satisfy the runtime's admissibility rules and asks only whether a safe opportunity is economically worth materializing. It is thus complementary to semantic recoverability~\citep{yang2026dart}, resume conformance~\citep{khan2026resume}, rollback defenses~\citep{zheng2026acrfence}, and semantic-environment isolation~\citep{mozafari2026semiso}.

\section{Measuring checkpoint advantage}
\paragraph{Recovery tracers.} Offline, we choose a candidate $s_t$ and a later failure $f$. We execute two matched worlds: \textsc{CP}, which materializes $s_t$ and recovers from it after $f$; and \textsc{Skip}, which recovers from $c^{-}$. Their measured cost difference estimates $A_{\mathrm{CP}}(s_t)$. Both branches restore into isolated workspaces, run the same pinned agent configuration against the same logical failure step, and stop under the same rule: success on the task's hidden-test verifier, or a predeclared recovery budget. Branch order is randomized per pair. Because the agent runtime exposes no controllable sampling seed, repetition rather than seed matching controls model nondeterminism; the recorded pair seed governs branch order and scheduling only. The label is thus produced by executed system behavior, with no LLM judge and no human annotation; Figure~\ref{fig:counterfactual-labeling} in Appendix~\ref{app:oracle} diagrams the construction.

\paragraph{Cost accounting.} The primary cost $J$ is wall-clock recovery time plus measured checkpoint creation time, so \textsc{CP} is charged its own overhead and a positive advantage is a net saving. Repeated tokens and tool calls, checkpoint bytes, restore latency, and task correctness are recorded separately, never folded into $J$.

\paragraph{From labels to placement (proposed, not evaluated here).} The labels are meant to train a placement policy from features cheap to compute online, such as execution cost, state size, verifier progress, and provenance-graph statistics, with a gradient-boosted regressor~\citep{chen2016xgboost} predicting $\widehat{A}_{\mathrm{CP}}(s_t)$ under a budget rule; Appendices~\ref{app:features}--\ref{app:baselines} specify it. This paper measures $A_{\mathrm{CP}}$ only, and Section~\ref{sec:pilot} reports how much room such a policy would have.

\section{Pilot: paired recovery on real agent executions}
\label{sec:pilot}

\paragraph{Setup.} We measure checkpoint advantage on \NumTasks{} from SWE-bench Verified~\citep{jimenez2024swebench}, four each from Django, SymPy, and scikit-learn, frozen before any paired run. A pinned agent (\PrimaryAgentModel) produced \NumTrajectories{} of \TrajStepsRange{} steps (mean \TrajStepsMean); \NominalSuccess{} passed the verifier. Checkpoints are \CheckpointBackendShort. From \NumNominalCandidates{} safe boundaries we froze \NumCandidates, namely \FailureWindow, with one failure each, then ran two repetitions per cell plus a predeclared third for the \StabilizedCells{} sign-unstable cells, for \NumCounterfactuals. Cells repeat one state, so every interval resamples tasks (\BootstrapReps{} replicates, seed \BootstrapSeed; Appendix~\ref{app:protocol}).

\paragraph{What the design contrasts.} Taking candidates in trajectory order fixes two things at once, and they must be read apart. In every early cell the candidate is the \emph{first} safe boundary, so \textsc{Skip} restores the pristine workspace and the contrast is one checkpoint versus none. In every late cell an earlier checkpoint exists, so \textsc{Skip} restores a mid-task state one step back, giving one \emph{additional} checkpoint versus the one already taken. Redo distance is identical in both cases, one step for \textsc{CP} and two for \textsc{Skip}, so the strata are separated by fallback identity rather than distance.

\begin{table}[t]
\centering
\small
\caption{Marginal value of a checkpoint, by what the \textsc{Skip} world falls back to. Each row is a within-cell paired contrast over the \VerTasks{} tasks whose recovery reached the endpoint verified by the task's hidden tests. Brackets are \BootstrapReps-replicate task-bootstrap 95\% intervals; advantage is net of measured creation cost, and conversion is advantage over the nominal work protected.}
\label{tab:marginal-value}
\begin{tabular}{lrrrrr}
\toprule
Contrast & Tasks & CP (s) & SKIP (s) & Marginal value (s) & Conversion \\
\midrule
First checkpoint vs.\ none & \VerTasks & \FirstCPCP & \FirstCPSKIP & \FirstCPAdvantage~[72.4, 131.1] & \FirstConversion \\
One additional, one step later & \VerTasks & \AddCPCP & \AddCPSKIP & \AddCPAdvantage~[$-9.5$, $8.4$] & \AddConversion \\
\midrule
Pooled over both strata & \VerTasks & \VerCP & \VerSKIP & \VerAdvantage~[34.1, 65.5] & \\
\bottomrule
\end{tabular}
\end{table}

\paragraph{The advantage is real, and it belongs to one regime.} Pooled over both strata, checkpointing saves \VerAdvantage\,s per task (\ConfidenceInterval{} \VerAdvantageCI), net of the \CheckpointOverheadMean\,s creation cost charged to \textsc{CP}. The decomposition in Table~\ref{tab:marginal-value} places almost all of it on the first checkpoint, worth \FirstCPAdvantage\,s (95\% CI \FirstCPAdvantageCI) and \FirstTokenSaving{} re-executed tokens. A second checkpoint one step later returns \AddCPAdvantage\,s (95\% CI \AddCPAdvantageCI), an interval that straddles zero and is smaller in magnitude than the cost of taking it. \textsc{CP} cost is nearly identical across strata, \FirstCPCP\,s against \AddCPCP\,s, while the late stratum preserves strictly more progress. The regimes differ by two orders of magnitude, so the decomposition rather than the pooled mean is what to design against.

\paragraph{Preserved work is a poor guide to saved work.} A checkpoint at candidate step $k$ protects exactly step~$k$'s nominal work, which makes the conversion from protected to saved work directly measurable. The first checkpoint protects \FirstProtected\,s and returns \FirstCPAdvantage\,s, a conversion of \FirstConversion. The second protects \AddProtected\,s and returns \AddCPAdvantage\,s, a conversion of \AddConversion. Work concentration accounts for part of the gap, since these trajectories place \StepOneShare{} of nominal work in step~1, but not for the collapse in conversion. An elapsed-work rule~\citep{young1974checkpoint} assumes conversion near one and prices the second checkpoint at its full \AddProtected\,s. Recovery is a re-derivation rather than a replay, and only a paired measurement separates the two.

\paragraph{Cheap proxies rank inside a regime, not across it.} Ordinary work proxies rank measured advantage well among the first-checkpoint cells and not at all among the additional-checkpoint cells (Table~\ref{tab:proxy}). Ordering candidates is therefore the easy part, and the signal a policy needs is the regime boundary itself. Two rows are worth naming. Bytes changed ranks at \ProxyBytes, so checkpoint value here is not sparse in bytes and the sparse-but-influential state our intuition describes does not arise in SWE-bench issue resolution. The importance score the agent assigns each step of its own accord, without being asked to price recovery, ranks \ProxyImportance{} and never positively in any subset, which is why labels come from executed behaviour rather than the model's judgement. The null is regime-specific: conditioned on the late failure alone the later boundary is cheaper on \LaterWinsLateFailure{} of \NumTasksN{} tasks.

\begin{table}[t]
\centering
\small
\caption{Spearman correlation between a cheap online proxy and measured advantage, within stratum over verified-endpoint cells; pooling the strata would manufacture a correlation from the split alone. $p$-values are permutation tests over \(20{,}000\) relabelings; the last column drops the largest cell.}
\label{tab:proxy}
\begin{tabular}{lrrr}
\toprule
Proxy & First checkpoint ($n{=}\VerTasks$) & Additional ($n{=}\VerTasks$) & Drop largest \\
\midrule
Tool calls & \ProxyTool~($p{=}\ProxyToolP$) & $+0.34$ & \ProxyToolDrop \\
Files modified & \ProxyFiles~($p{=}\ProxyFilesP$) & $-0.53$ & $+0.72$ \\
Bytes changed & \ProxyBytes~($p{=}\ProxyBytesP$) & $-0.29$ & \ProxyBytesDrop \\
Agent importance & \ProxyImportance~($p{=}\ProxyImportanceP$) & $+0.06$ & $-0.49$ \\
\bottomrule
\end{tabular}
\end{table}

\section{Related work}
Agent checkpoint and restore is moving quickly: \textsc{Crab} on semantics-aware checkpoint/restore~\citep{wu2026crab}, \textsc{DeltaBox} on low-latency sandbox C/R~\citep{dong2026deltabox}, \textsc{DART} on admissible local recovery~\citep{yang2026dart}, plus resume contracts~\citep{khan2026resume}, rollback safety~\citep{zheng2026acrfence}, causal progress~\citep{wu2026structagent}, and semantic isolation~\citep{mozafari2026semiso}. \RippleCP{} claims no new mechanism; it asks what a safe opportunity those systems expose is economically worth.

That question has a long lineage outside agents. Adjoint checkpointing places a bounded number of snapshots along a computation to trade storage against recomputation~\citep{griewank2000revolve}, activation checkpointing makes the same trade inside a training step~\citep{chen2016sublinear}, materialized view selection persists intermediate results under a space budget~\citep{harinarayan1996datacubes}, and classical interval placement balances overhead against expected redo~\citep{young1974checkpoint}. In all of these, recomputation cost is deterministic and analytically known, so placement reduces to optimization. For a stochastic agent it is neither. Recovery is a re-derivation rather than a replay, so its cost must be measured, and Section~\ref{sec:pilot} shows the measured cost and the analytic proxy differing by the whole quantity.

Causal Agent Replay~\citep{shah2026causalreplay} intervenes on real LLM-agent trajectories to attribute failures to individual steps, and meets the same obstacle we do: under a stochastic policy, re-executing a trajectory perturbs everything downstream of the intervention. It resolves this with a point-of-commitment rule over resampled rollouts, whereas we use repeated paired branches, randomized order, and a predeclared sign-stability criterion. The questions differ, one asking which step caused a failure and the other what a checkpoint there is worth, but the instrumentation problem is shared and neither yields to replaying recorded costs.

\section{Discussion and limitations}
\label{sec:limitations}
Three things follow for a runtime builder. Take one checkpoint at the first safe boundary: it captures the available value for \CheckpointOverheadMean\,s. Do not spend budget ranking later boundaries until the workload is known to have more than one expensive step. Price a candidate by the recovery it avoids rather than the work it preserves. The bar for a learned policy is now concrete: beat the first-boundary rule at a matched budget on a workload where conversion varies across candidates.

That bar is not clearable here, which is the main limitation. Trajectories are \TrajStepsRange{} steps (mean \TrajStepsMean) with \StepOneShare{} of the work in the first, so the workload is close to a single expensive step followed by verification. Re-cutting the same tasks at tool-call boundaries would yield \RecutPerTask{} candidates per task, \RecutTurnOne{} inside that first step, though the logs carry no per-item timings and so need re-collection rather than re-analysis (Appendix~\ref{app:results}). Workloads with heterogeneous accumulated state, such as dependency resolution or multi-stage builds, are where the hypothesis lives.

Three constraints bound the claims. The first safe boundary is always the one whose \textsc{Skip} world restarts from scratch, so ``first'' and ``earliest'' are not separable. The additional-checkpoint estimate sits at per-cell resolution, \NoiseAdd\,s of noise against a \NoiseAddEffect\,s effect, so the task-level bootstrap rather than any single cell supports it. And \NumTasksN{} tasks from three repositories in one tier, one agent, one model version is a small effective sample. We report no learned predictor and no matched-budget study; Appendices~\ref{app:training} and~\ref{app:baselines} specify both.

\section{Conclusion}
Checkpoint placement for agents is a question about value, not mechanism. Checkpointing saves \VerAdvantage\,s per task pooled, but \FirstCPAdvantage\,s of that belongs to the first checkpoint and a second one step later returns \AddCPAdvantage\,s on \AddProtected\,s of protected work. The quantity that governs placement is the conversion from preserved work to saved work, which this harness measures and elapsed-work proxies assume.

\bibliographystyle{unsrtnat}
\bibliography{references}

\clearpage
\appendix
\section{Detailed experimental protocol}
\label{app:protocol}
\paragraph{Task selection.} Task IDs were frozen before any paired run. The workload is \NumTasks{} from SWE-bench Verified, the human-validated subset of SWE-bench~\citep{jimenez2024swebench}: four each from \texttt{django/django}~4.2, \texttt{sympy/sympy}~1.12, and \texttt{scikit-learn/scikit-learn}~1.3, all drawn from the benchmark's 15\,min--1\,hour difficulty tier. Exact issue IDs, base commits, benchmark source checksum, hidden-test patches, normalized verifier commands, and pinned Python 3.10 environments are retained with the run artifacts. Base-fail/gold-pass validation succeeded for all \NumTasksN{} tasks and all 24 expected worlds; this establishes benchmark integrity only and is not an agent result. Tasks that never reach a safe candidate would be excluded and reported; none were. Because this study measures $A_{\mathrm{CP}}$ rather than fitting a predictor, no train/validation/test split was drawn, and uncertainty is a task-level bootstrap throughout. A predictor study would additionally require repository-level separation, ideally leave-one-repository-out.

\paragraph{Trajectory collection.} One nominal trajectory per task was recorded with \PrimaryAgentModel{} at sandbox level \texttt{workspace-write} and approval policy \texttt{never}, capturing every model call and observable tool boundary with timestamps, token counts (input, cached input, output), tool arguments and return status, workspace hashes, verifier outcomes, and produced artifacts. Raw JSONL event streams, stderr, prompts, verifier outputs, and per-branch timings are retained beneath the run directory. The agent runtime does not expose a controllable model-sampling seed; the recorded pair seed therefore controls branch order and experiment scheduling, not model randomness. This is a limitation of the harness, not a design choice, and is the reason repetitions rather than seed matching are used to control nondeterminism.

\paragraph{Candidate boundaries.} Candidates come only from safe boundaries exposed by the runtime: post-tool-return, post-commit, and post-test points. The placement model never decides checkpoint correctness. From \NumNominalCandidates{} nominal safe boundaries we selected \NumCandidatesN{} cells under a frozen stratification: \FailureWindow. Candidates therefore sit at steps \CandidateSteps{} and failures at steps \FailureSteps. Selection was completed before any paired recovery was executed. Because candidates are taken in trajectory order, the first selected boundary is also the only one whose \textsc{Skip} world has no earlier checkpoint to fall back on: the early cells restore the pristine workspace and the late cells restore a mid-task state one logical step back. Boundary position and fallback identity are therefore coupled by construction, which is why Section~\ref{sec:pilot} reports the strata as two marginal contrasts rather than as an comparison of early against late. This realizes the temporal axis of the protocol only. The protocol also called for stratifying over physical size, semantic progress, and ripple score, and for including late as well as near failure offsets; at this scale we spent the cell budget on the timing contrast and on repetition instead. One consequence is that the pilot cannot test the paper's motivating ``sparse in bytes, dense in downstream influence'' hypothesis, which needs the size and ripple axes.

\paragraph{Paired execution and stopping rule.} Each cell was run with two repetitions by default. A separate, resumable stabilization pass then froze the cells whose first two paired advantages changed sign and ran exactly one additional pair for each, adding \StabilizedCells{} pairs and preserving the predeclared 2-or-3 repetition rule without broadening expense. Branch order was randomized per pair (26 \textsc{CP}-first, 27 \textsc{Skip}-first across \NumPairs{} pairs). Both branches of a pair share the same task state, logical failure step, model and tool settings, and stopping rule. A branch runs the agent to its own completion under a hard per-branch cap of \AgentTimeoutCap\,s, after which the task's hidden-test verifier gates success. The cap was never reached, since the longest branch spent \LongestBranch\,s of agent time, so in practice every branch terminated on agent completion and was then judged by the verifier. Branches on the two tasks whose nominal trajectory never passed the verifier therefore terminate with a failing verdict rather than at a cost ceiling, which is why those tasks enter the analysis only as a termination-cost sensitivity.

\paragraph{Trustworthy and adaptive foundation-model systems.}
More broadly, reliable agent execution depends not only on recovering state, but also on controlling what models retain, adapt, expose, and act upon: recent work studies selective unlearning across LLMs, vision models, and embodied agents~\cite{ranjan2026forgetting,ranjan2026razor,ranjan2026vla}, together with deployment constraints for embodied foundation models at the edge~\cite{grover2026embodied}. 
Complementary directions address privacy leakage~\cite{ranjan2026g}, fairness and retrieval-guided debiasing~\cite{ranjan2026catrag,ranjan2026position}, personalized adaptation~\cite{ranjan2026persa}, and model explainability~\cite{ranjan2026listening}, highlighting that trustworthy behavior depends on preserving or modifying the right internal and external information at the right time. 
For memory-grounded agents, SafeCommit further studies when accumulated state is sufficient for an agent to act safely~\cite{akewar2026safecommit}; RippleCP is complementary in asking when such recoverable agent state is economically worth materializing as a checkpoint.

\paragraph{Checkpoint contents.} A checkpoint is \CheckpointBackend. Concretely, the backend archives the task workspace, excluding \texttt{.git}, virtualenvs, and build/cache directories, together with the preserved transcript and discovered facts, and records creation latency and compressed size for every artifact. Checkpoint creation latency was \CheckpointLatency{} over the \NumPairs{} \textsc{CP} branches; restore latency was \RestoreLatency{} over all \NumBranches{} branches; compressed checkpoint size was \CheckpointBytes{} over the 39 stored artifacts. Process state and tool-session state are not captured; the harness reconstructs agent-visible state from the archived transcript rather than from an OS-level image.

\section{Counterfactual advantage oracle}
\label{app:oracle}
For each trajectory, candidate $s_t$, and sampled future failure $f_j$:
\begin{enumerate}[leftmargin=*]
  \item restore to $s_t$ from a common pre-candidate state;
  \item in the \textsc{CP} branch, materialize $s_t$; in \textsc{Skip}, do not;
  \item drive both branches to the same logical failure location and inject a fail-stop termination;
  \item recover \textsc{CP} from $s_t$ and \textsc{Skip} from $c^{-}$ under matched model, tool, and verifier settings, in randomized branch order;
  \item continue to the task's verified endpoint or a fixed recovery budget, applying the same stopping rule to both worlds;
  \item record total redo time, repeated tokens and tool calls, task correctness under the hidden-test verifier, and measured C/R overhead.
\end{enumerate}
Estimate
\begin{equation}
\widehat{A}_{\mathrm{CP}}(s_t)=\frac{1}{K}\sum_{j=1}^{K}\big[J_j(s_t,\mathrm{SKIP})-J_j(s_t,\mathrm{CP})\big].
\end{equation}
In the pilot of Section~\ref{sec:pilot}, $K=1$: each selected candidate carries a single frozen logical failure, and repetition is spent on re-running that same cell rather than on sampling additional offsets. This is a deliberate trade at pilot scale. It buys tighter per-cell estimates under a stochastic agent at the cost of averaging over the failure distribution, and it is why the analysis stratifies over candidates by what the \textsc{Skip} world falls back to, rather than sweeping offsets within a candidate. A study that aims to characterize $\mathcal{F}$ itself should invert that trade.

\begin{figure}[h]
\centering
\resizebox{0.70\linewidth}{!}{%
\begin{tikzpicture}[
    font=\small,
    >=Latex,
    box/.style={
        draw,
        rounded corners=2pt,
        align=center,
        minimum height=8mm,
        text width=2.35cm,
        inner sep=3pt
    },
    smallbox/.style={
        draw,
        rounded corners=2pt,
        align=center,
        minimum height=7mm,
        text width=1.9cm,
        inner sep=3pt
    },
    costbox/.style={
        draw,
        rounded corners=2pt,
        align=center,
        minimum height=9mm,
        text width=1.80cm,
        inner sep=3pt,
        font=\small\bfseries
    },
    labelbox/.style={
        draw,
        rounded corners=3pt,
        align=center,
        text width=4.7cm,
        inner sep=4pt
    },
    arr/.style={
        ->,
        thick
    },
    dashedarr/.style={
        ->,
        thick,
        dashed
    }
]


\node[box] (prev) at (0,0)
{\textbf{Previous checkpoint}\\
$c^{-}$};

\node[box] (cand) at (3.0,0)
{\textbf{Candidate state}\\
$s_t$};

\node[smallbox] (cp) at (6.2,1.35)
{\textbf{CP world}\\
checkpoint $s_t$};

\node[smallbox] (skip) at (6.2,-1.35)
{\textbf{SKIP world}\\
do not checkpoint};

\node[smallbox] (failcp) at (9.15,1.35)
{\textbf{Same future failure}\\
at time $t+\Delta$};

\node[smallbox] (failskip) at (9.15,-1.35)
{\textbf{Same future failure}\\
at time $t+\Delta$};

\node[costbox] (jcp) at (13.75,1.35)
{$J(\mathrm{CP})$\\
recovery cost};

\node[costbox] (jskip) at (13.75,-1.35)
{$J(\mathrm{SKIP})$\\
recovery cost};

\draw[arr] (prev) -- (cand);

\draw[arr] (cand.east) -- ++(0.55,0) |- node[pos=0.30,right,font=\scriptsize]{branch} (cp.west);
\draw[arr] (cand.east) -- ++(0.55,0) |- (skip.west);

\draw[arr] (cp) -- (failcp);
\draw[arr] (skip) -- (failskip);

\draw[arr] (failcp) -- node[above,font=\scriptsize,inner sep=1pt]{recover from $s_t$} (jcp);
\draw[arr] (failskip) -- node[below,font=\scriptsize,inner sep=1pt]{recover from $c^{-}$} (jskip);


\node[labelbox] (adv) at (5,-3.0)
{
\textbf{Checkpoint Advantage label}\\[2pt]
$A_{\mathrm{CP}}(s_t) = J(\mathrm{SKIP}) - J(\mathrm{CP})$\\[2pt]
Positive value $\Rightarrow$ checkpointing $s_t$ avoids future recovery cost.
};

\draw[dashedarr]
    (jcp.east) -- ++(0.65,0)
    |- ([yshift=5pt]adv.east);

\draw[dashedarr]
    (jskip.east) -- ++(0.40,0)
    |- ([yshift=-5pt]adv.east);

\node[font=\scriptsize, align=center] at (3.0,1.0)
{Evaluate whether saving\\this state is worthwhile};

\node[font=\scriptsize, align=center] at (13.75,2.55)
{Measure cost in time,\\tokens, tools, or total failure cost};

\end{tikzpicture}%
}
\caption{\textbf{Counterfactual labeling.} From the previous checkpoint $c^{-}$ and candidate $s_t$ we run two matched worlds, \emph{CP}, which materializes $s_t$, and \emph{SKIP}, which does not, to the same future failure, then recover both and measure $J(\mathrm{CP})$ and $J(\mathrm{SKIP})$. Their difference is the label $A_{\mathrm{CP}}(s_t)$.}
\label{fig:counterfactual-labeling}
\end{figure}
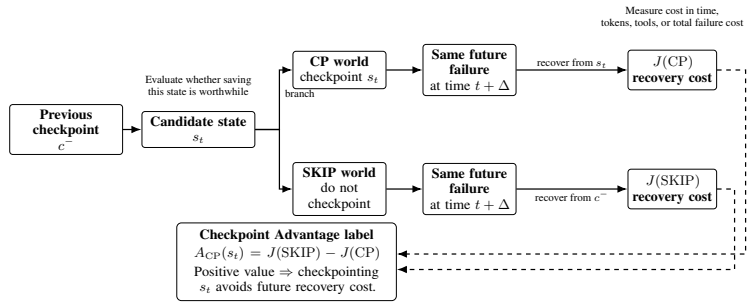

The oracle is an expected systems measurement, not an LLM judgment. Because the agent is stochastic and no sampling seed is controllable, individual pair differences can change sign; the protocol handles this by repeating unstable cells and by resampling tasks for uncertainty, never by discarding a sign-inconsistent observation.

\section{Feature and ripple-graph specification}
\label{app:features}
This appendix specifies the feature families intended for a learned placement policy. None of them is evaluated in this paper: the pilot measures $A_{\mathrm{CP}}$ and does not fit a predictor, so no correlation between any feature below and measured advantage is reported or implied.

\begin{table}[h]
\centering
\small
\caption{Proposed online feature families for advantage prediction. Not evaluated in this paper.}
\begin{tabular}{p{0.20\linewidth}p{0.70\linewidth}}
\toprule
Group & Examples \\
\midrule
Execution cost & elapsed seconds; tokens since last checkpoint; tool calls; retries; errors; expensive-tool count \\
Physical state & files touched; bytes modified; artifact size; optional process-state size \\
Verified progress & newly passing tests; completed subgoals; verifier events; newly validated artifacts/facts \\
Ripple / graph & new descendant count; dependency fanout; downstream verified nodes; affected artifact count; centrality proxy \\
Context & distance from last checkpoint; time since last progress; branch depth; recent error trend \\
\bottomrule
\end{tabular}
\end{table}

Edges must be observable: command$\rightarrow$file-write, file-version$\rightarrow$test-result, build-input$\rightarrow$build-output, tool-result$\rightarrow$artifact, or explicit workflow dependencies. If any natural-language dependency is inferred by an LLM, it should be isolated as a separate feature family and ablated. Calling these dependency or ripple features is a claim about observable provenance, not about causal influence; the counterfactual recovery measurement, not the graph, is the ground-truth target.

\section{Small-data training protocol (future work)}
\label{app:training}
No predictor is trained in this paper. This appendix records the protocol a follow-on study should use, so that the measurement design and the eventual learning design can be reviewed together.

The project assumes no pre-existing checkpoint-value dataset; labels come from the counterfactual oracle of Appendix~\ref{app:oracle}. Statistical independence is at the task/repository level, so candidate states must never be split randomly: all states from one task belong to one partition, and leave-one-repository-out is preferable once enough repositories are available. The pilot reported here yields \NumCandidatesN{} labelled cells from \NumTasksN{} tasks, which is enough to estimate the population effect but not enough to fit and honestly evaluate a predictor; scaling the labelled set is the precondition for the rest of this appendix.

\paragraph{Primary model.} Start with linear/ridge regression and gradient-boosted trees. The target is measured $A_{\mathrm{CP}}(s_t)$; an alternative binary target is $\mathbb{1}[A_{\mathrm{CP}}>0]$, which the pilot suggests is a meaningfully unbalanced problem (\PositiveCells{} of \NumCandidatesN{} cells positive). Tune only on validation tasks. Report rank quality alongside error, because placement is a prioritization problem.

\paragraph{Required baseline.} Any predictor must be compared against the rule \emph{implied by} the measured decomposition, namely take the first checkpoint then stop, at a matched budget. The pilot measures the decomposition; it does not run such a policy. Given how sharply advantage separates by fallback identity (Section~\ref{sec:pilot}), a learned model that does not beat that rule has not demonstrated that learning is necessary, and the correct conclusion would be to report the negative result and keep the simpler rule.

\paragraph{Why not deep RL first?} Online agent episodes are expensive and the action space is only \{checkpoint, skip\}. The supervised advantage label already estimates the local action difference directly. Deep Q-learning is justified only if sequential interaction among checkpoint decisions materially improves end-to-end cost; Appendix~\ref{app:qrl} states that hypothesis separately.

\section{Optional sequential Q-learning extension (future work)}
\label{app:qrl}
Checkpoint placement is sequential because a checkpoint changes the identity of the next previous checkpoint and consumes budget. To test whether this interaction matters, define simulator state
\[
z_t=[x_t,\;\text{distance to last CP},\;B_{\mathrm{remaining}}]
\]
and actions $a_t\in\{\mathrm{CP},\mathrm{SKIP}\}$. Rewards are negative checkpoint cost when materializing and negative redo/recovery cost when a simulated failure occurs. A fitted-Q iteration (FQI) baseline~\citep{ernst2005fqi} would learn $Q(z,a)$ from trace-replay transitions rather than from expensive new online agent episodes.

\paragraph{Decision criterion.} \QRipple{} should enter the main paper only if it provides a meaningful held-out benefit over local advantage ranking at the same budget. If the gain is negligible, the negative result should be reported and the simpler model kept. This prevents ``RL for RL's sake.'' Neither \QRipple{} nor the local ranking policy it would be compared against has been run on the real workload, so this paper reports no result for either.

\paragraph{A caution from the pilot.} Trace-replay transitions are only as good as the cost model behind them. The additional-checkpoint contrast shows a mean advantage of \AddCPAdvantage\,s, smaller in magnitude than the \CheckpointOverheadMean\,s cost of taking the checkpoint, so a simulator that models redo cost as monotone in elapsed work would systematically misprice exactly the decisions a budgeted policy has to get right. Any fitted-Q study should therefore be validated against measured paired recoveries, not only against replayed traces.

\section{Baselines, budgets, and metrics (specification for the policy study)}
\label{app:baselines}
The baselines below are specified so that the eventual policy comparison is fixed in advance. None was run in this paper, and no claim is made about how \RippleCP{} would compare to any of them.

\paragraph{Baselines.}
\begin{description}[leftmargin=!,labelwidth=1.20in]
  \item[No checkpoint] Restart from task beginning.
  \item[Every turn] Upper-overhead reference; not a budget-matched competitor.
  \item[Fixed-$k$] Checkpoint every $k$ safe boundaries; tune $k$ to match budget.
  \item[Time-based] Fixed wall-clock interval, validation-tuned to match budget.
  \item[First-boundary] Spend the budget on the earliest safe boundaries. This is the rule implied by the decomposition in Section~\ref{sec:pilot} and is the baseline a learned policy most needs to beat.
  \item[State-change] Trigger by bytes/files changed using a validation threshold.
  \item[Progress] Trigger after a predefined verifier/test/subgoal progress event.
  \item[Failure-risk] Predict near-term failure using non-ripple features and preserve highest-risk states. This separates ``failure is likely'' from ``state is valuable.''
  \item[Random] Uniform safe candidates at the same budget; average multiple seeds.
  \item[Oracle] Rank by measured $A_{\mathrm{CP}}$; unattainable upper bound.
  \item[Q-Ripple] Optional fitted-Q policy from Appendix~\ref{app:qrl}.
\end{description}

\paragraph{Primary metrics.}
\begin{align*}
\mathrm{LostWork} &= \text{wall-clock re-execution after injected failure},\\
\mathrm{RecoveryEfficiency} &= \frac{\text{redo time avoided vs. no checkpoint}}{\text{checkpoint time overhead}},\\
\mathrm{TotalFailureCost} &= \text{normal run cost}+\text{C/R overhead}+\text{redo cost}.
\end{align*}
Report task completion, repeated tokens and tool calls, bytes written, checkpoint latency, and restore latency separately rather than folding them into one weighted scalar.

\paragraph{Matched budgets.} Compare policies at equal checkpoint count and, where C/R cost varies by checkpoint size, at equal measured checkpoint-time or checkpoint-byte budgets. A placement method must not win merely by checkpointing more often. The pilot supplies the overhead constants such a comparison needs: \CheckpointLatencyMean\,s mean creation cost and \CheckpointBytes.

\section{Extended pilot results}
\label{app:results}
This appendix reports the full pilot results underlying Section~\ref{sec:pilot}. Every value is computed from the retained run records by the analysis code released with the paper, and the two analysis paths that overlap are checked against each other and agree exactly. Appendices~\ref{app:training} and~\ref{app:baselines} state what each would require.

\subsection{All-task termination-cost sensitivity}
The main-text estimates are restricted to the \VerTasks{} tasks where both branches reached the endpoint verified by the hidden tests. Including the two tasks that neither branch solved, the pooled figure becomes \AllAdvantage\,s (95\% CI \AllAdvantageCI) over all \NumTasksN{} frozen tasks, with \textsc{CP} at \AllCP\,s and \textsc{Skip} at \AllSKIP\,s. We report this only as a sensitivity: for the unsolved tasks it measures cost under the fixed recovery procedure, not time to a verified fix, so it is not comparable to the verified-endpoint numbers and is not used for any claim.

\subsection{Marginal value of a checkpoint}
\begin{figure}[h]
  \centering
  \includegraphics[width=0.78\linewidth]{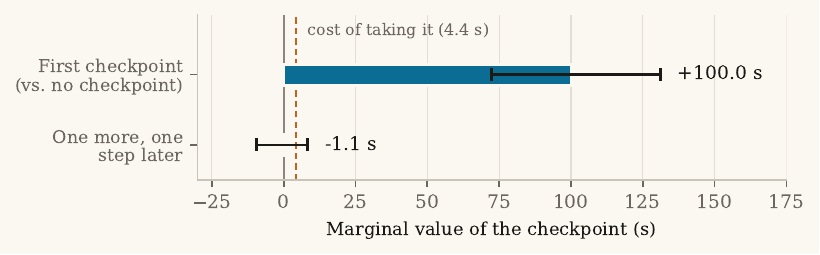}
  \caption{Marginal value of a checkpoint by what the \textsc{Skip} world falls back to, over the \VerTasks{} verified-endpoint tasks, with task-bootstrap 95\% intervals. The dashed rule is measured creation cost. These are the first two rows of Table~\ref{tab:marginal-value}.}
  \label{fig:marginal-value}
\end{figure}

\subsection{Do cheap proxies rank measured advantage?}
Table~\ref{tab:proxy} in the main text reports these correlations, including the discovered-facts row omitted there for space (\ProxyFacts, $p{=}\ProxyFactsP$, and $-0.21$ in the additional-checkpoint cells). Rank correlations are computed \emph{within} stratum. The two contrasts differ by two orders of magnitude, so pooling them would manufacture a correlation out of the stratum split alone. $p$-values are two-sided permutation tests over \(20{,}000\) relabelings, which is the right tool at this $n$; the normal approximation would overstate confidence.

Three readings. Work proxies rank advantage well inside the first-checkpoint regime and not at all inside the additional-checkpoint regime, so the difficulty in placement is identifying which regime a candidate is in, not ordering candidates once the regime is known. Bytes changed ranks at \ProxyBytes, which locates the motivating intuition rather than confirming it: on this workload checkpoint value is \emph{not} sparse in bytes, and the contrast behind \RippleCP{} (Appendix~\ref{app:intuition}) belongs to workloads that accumulate heterogeneous state. The importance score the agent assigns each step of its own accord, without being asked to price recovery, never ranks positively in any subset, which at $n{=}\VerTasks$ is suggestive rather than conclusive and is why labels come from executed behaviour instead of the model's judgement.

\paragraph{Why checkpoint archive size is not in that table.} Archive size correlates $-0.31$ with advantage, which invites reading as evidence that physical size does not predict value. It is not. Archive size here is a repository fingerprint: within-repository spread is \ArchiveDjango{} for Django, \ArchiveSklearn{} for scikit-learn, and \ArchiveSympy{} for SymPy, against \ArchiveOverall\,MB of overall spread. Correlating it with advantage largely tests which repository a cell came from. A genuine size feature has to vary within a repository, which is what the bytes a step modifies does.

\subsection{Measurement reliability}
Mean within-cell standard deviation of paired advantage across repetitions is \NoiseAll\,s. Split by stratum, the first-checkpoint cells have \NoiseFirst\,s of noise against a \NoiseFirstEffect\,s mean effect (noise-to-signal \NoiseFirstRatio), while the additional-checkpoint cells have \NoiseAdd\,s against \NoiseAddEffect\,s (noise-to-signal \NoiseAddRatio). The additional-checkpoint effect is therefore at the resolution of a single cell measured two or three times, and the aggregate claim rests on the task-level bootstrap rather than on any individual cell. A study wanting to resolve an effect of that size per cell should budget roughly an order of magnitude more repetitions. For planning, one label cost \CostPerLabel\,min of agent time.

\subsection{What a finer candidate set would cost}
The frozen study takes candidates at coarse turn boundaries: one early and one late per task. The nominal event logs already record every tool call, so the boundaries for a finer re-cut can be counted without collecting anything new. Across the \NumTasksN{} nominal trajectories there are \RecutItems{} completed tool-call items (\RecutCommands{} command executions, \RecutFileChanges{} file changes, the rest agent messages and planning items): \RecutPerTask{} per task, of which \RecutTurnOne{} on average (range \RecutTurnOneRange) fall inside the first turn, the one carrying \StepOneShare{} of the nominal work, where this design currently places exactly one candidate.

That is the missing resolution, and it is available on the existing workload. What is \emph{not} available is the cost of those boundaries: the recorded events carry turn-level token and timing totals only, with no per-item timestamps, so the work protected by a tool-call candidate cannot be priced from the current artifacts. A re-cut therefore requires re-collecting nominal trajectories with per-item timing instrumented, after which the paired-recovery harness applies unchanged. We flag this as the concrete next step rather than an open-ended call for a deeper workload.

\subsection{Where the nominal work is}
\begin{table}[h]
\centering
\small
\caption{Distribution of nominal (pre-failure) work over logical steps, across all \NumTasksN{} frozen trajectories. A checkpoint at candidate step $k$ protects exactly step $k$'s work: the \textsc{Skip} world restores at $k-1$ and re-executes step $k$ before continuing to the same failure. The workload is close to a single expensive step followed by verification.}
\label{tab:work-concentration}
\begin{tabular}{lrrrr}
\toprule
Step & Trajectories & Mean (s) & Share & Cumulative \\
\midrule
1 & 12 & \StepOneMean & \StepOneShare & \StepOneShare \\
2 & 12 & \StepTwoMean & \StepTwoShare & \CumTwo \\
3 & 12 & \StepThreeMean & \StepThreeShare & \CumThree \\
4 & 3 & \StepFourMean & \StepFourShare & 100\% \\
\bottomrule
\end{tabular}
\end{table}

Averaged per task rather than pooled, the first step carries \StepOneShareMean{} of that task's nominal work, and it exceeds 70\% on 9 of \NumTasksN{} tasks. Matching each stratum's candidate step to that step's nominal duration on the \VerTasks{} verified-endpoint tasks, the first checkpoint protects \FirstProtected\,s of nominal work and returns \FirstCPAdvantage\,s of measured advantage (conversion \FirstConversion), while the additional checkpoint protects \AddProtected\,s and returns \AddCPAdvantage\,s (conversion \AddConversion). An elapsed-work placement rule, which implicitly assumes a conversion near one, would therefore have valued the second checkpoint at roughly \AddProtected\,s and been wrong by its full nominal cost. We report this as a two-point observation: the strata differ in fallback identity as well as in protected work, so it motivates advantage-based placement without constituting a controlled test of it.

\subsection{Recovery cost by restore depth}
\begin{table}[h]
\centering
\small
\caption{Mean recovery cost by how much progress the restore point preserved, over all \NumBranches{} branches. Descriptive only: these are not within-task paired contrasts, and the deepest bucket is drawn from the three four-step tasks alone. The paired contrasts in Table~\ref{tab:marginal-value} are the controlled version of the same comparison.}
\label{tab:restore-depth}
\begin{tabular}{lrrr}
\toprule
Restore point & Branches & Cost (s) & Tokens re-executed \\
\midrule
Pristine (start over) & 24 & \RestorePristine & 640{,}651 \\
After step 1 & 45 & \RestoreOne & 162{,}656 \\
After step 2 & 29 & \RestoreTwo & 154{,}439 \\
After step 3 & 8 & \RestoreThree & 118{,}976 \\
\bottomrule
\end{tabular}
\end{table}

\subsection{Retrospective one-checkpoint placement}
For a task whose nominal trajectory is exactly three logical steps, all four recoveries a single-checkpoint policy needs were executed: checkpointing the first boundary gives restore@1 against the step-2 failure (the early cell's \textsc{CP} branch) and restore@1 against the step-3 failure (the late cell's \textsc{Skip} branch); checkpointing the second boundary gives restore from pristine against the step-2 failure, because that checkpoint does not exist yet (the early cell's \textsc{Skip} branch), and restore@2 against the step-3 failure (the late cell's \textsc{CP} branch). We weight the two failure points equally. Four-step tasks are excluded because their late cell falls back to step 2, so ``checkpoint first, fail last'' was never run; a no-checkpoint policy is excluded because restore from pristine was only ever executed against the step-2 failure. This reuses frozen measurements and is not a matched-budget experiment.

\begin{table}[h]
\centering
\small
\caption{Retrospective single-checkpoint placement on the \BudgetOneAllTasks{} three-step tasks (\BudgetOneTasks{} of which reached the verified endpoint), under a uniform prior over the two measured failure points. Lower is better. An oracle that picks the better boundary per task coincides with the first-boundary row on \BudgetOneAllTasks/\BudgetOneAllTasks{} tasks, so it is not shown separately.}
\label{tab:budget-one}
\begin{tabular}{lrr}
\toprule
Single-checkpoint policy & Verified (s) & All (s) \\
\midrule
Checkpoint at first safe boundary & \textbf{\BudgetOneFirst} & \textbf{\BudgetOneFirstAll} \\
Checkpoint at second safe boundary & \BudgetOneSecond & \BudgetOneSecondAll \\
\bottomrule
\end{tabular}
\end{table}

\paragraph{This table adds no evidence beyond Table~\ref{tab:marginal-value}.} Writing $F$ and $S$ for the first- and second-boundary costs, $F=\tfrac{1}{2}(\textsc{cp}_{\text{early}}+\textsc{skip}_{\text{late}})$ and $S=\tfrac{1}{2}(\textsc{skip}_{\text{early}}+\textsc{cp}_{\text{late}})$, so $S-F$ is identically half the difference between the first-checkpoint and additional-checkpoint contrasts. The recast is useful because it states the decomposition in the units a system designer budgets in, not because it is an independent measurement, and its margin is dominated by the arm in which the second-boundary policy holds no checkpoint at all.

\begin{figure}[h]
  \centering
  \includegraphics[width=0.82\linewidth]{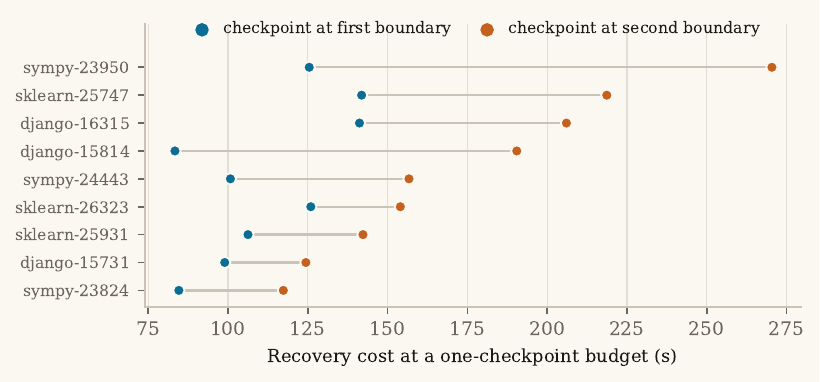}
  \caption{Retrospective one-checkpoint budget, per task, on the \BudgetOneAllTasks{} three-step tasks. Placing the single checkpoint at the first safe boundary is cheaper on every task. Because the second-boundary policy holds no checkpoint at the step-2 failure, this margin restates the first-checkpoint contrast rather than measuring placement.}
  \label{fig:budget-one}
\end{figure}

\paragraph{The one contrast that isolates placement.} On the \MidTasks{} four-step tasks the late cell sits at boundary three with its fallback at boundary two, so comparing ``checkpoint at boundary two'' against ``checkpoint at boundary three'' holds the early arm identical, since both restore from pristine at the step-2 failure, and varies only placement among mid-trajectory boundaries. The earlier boundary wins on \MidTasks/\MidTasks{} tasks, by \MidGap\,s (\MidBoundaryTwo\,s versus \MidBoundaryThree\,s). With \MidTasks{} tasks we quote no interval; the point is the order of magnitude, which is comparable to the \CheckpointOverheadMean\,s cost of the checkpoint rather than to the \FirstCPAdvantage\,s first-checkpoint effect. Conditioned on the late failure alone, the later boundary is in fact cheaper on \LaterWinsLateFailure{} of the \NumTasksN{} tasks, so a policy that knew where the failure would land could beat the fixed rule.

\subsection{Per-task advantage}
\begin{figure}[h]
  \centering
  \includegraphics[width=0.86\linewidth]{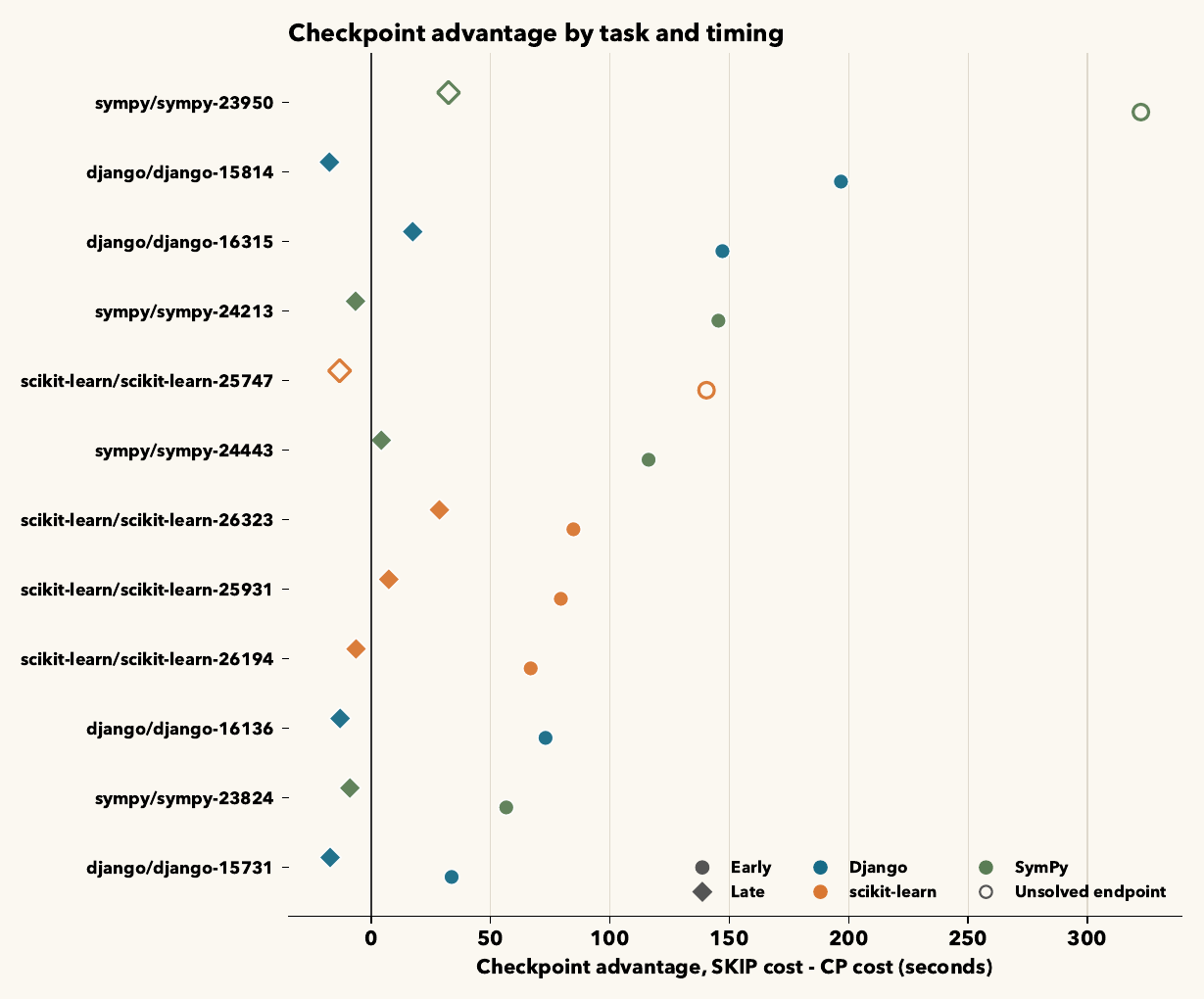}
  \caption{Per-task, per-stratum checkpoint advantage ($J(\mathrm{SKIP})-J(\mathrm{CP})$), by repository and stratum. Hollow markers are the two tasks whose nominal trajectory never reached the endpoint verified by the hidden tests; they are excluded from the primary estimate and shown here for completeness. The legend's ``Early'' and ``Late'' are the first-checkpoint and additional-checkpoint contrasts of Table~\ref{tab:marginal-value}. Additional-checkpoint points sit near zero in every repository (Django mean $-7.5$\,s, scikit-learn $+4.2$\,s, SymPy $+5.4$\,s), and 5 of the 12 are individually positive.}
  \label{fig:advantage-by-task}
\end{figure}

\begin{table}[h]
\centering
\small
\caption{Per-task means over the task's two frozen cells. Advantage is the mean of $J(\mathrm{SKIP})-J(\mathrm{CP})$ across those cells; positive favors checkpointing. ``Verified'' indicates whether both recovery branches reached the endpoint verified by the hidden tests (identical to the nominal outcome on this workload). ``Pairs'' counts repeated CP/SKIP measurements and sums to \NumPairs{} across the table.}
\label{tab:per-task}
\begin{tabular}{llrrrrr}
\toprule
Task & Repository & Verified & Pairs & CP (s) & SKIP (s) & Adv.\ (s) \\
\midrule
django-15731 & django/django & yes & 5 & 107.5 & 115.9 & 8.3 \\
django-15814 & django/django & yes & 4 & 92.1 & 181.8 & 89.7 \\
django-16136 & django/django & yes & 5 & 88.0 & 118.1 & 30.1 \\
django-16315 & django/django & yes & 4 & 132.5 & 214.8 & 82.3 \\
sklearn-25747 & scikit-learn & no & 4 & 148.4 & 212.1 & 63.7 \\
sklearn-25931 & scikit-learn & yes & 4 & 102.6 & 146.0 & 43.5 \\
sklearn-26194 & scikit-learn & yes & 5 & 104.8 & 135.0 & 30.3 \\
sklearn-26323 & scikit-learn & yes & 4 & 111.6 & 168.3 & 56.7 \\
sympy-23824 & sympy/sympy & yes & 5 & 89.0 & 112.9 & 23.9 \\
sympy-23950 & sympy/sympy & no & 5 & 109.2 & 286.6 & 177.4 \\
sympy-24213 & sympy/sympy & yes & 4 & 80.5 & 150.0 & 69.5 \\
sympy-24443 & sympy/sympy & yes & 4 & 98.6 & 158.9 & 60.3 \\
\bottomrule
\end{tabular}
\end{table}

\subsection{Per-repository breakdown}
\begin{table}[h]
\centering
\small
\caption{Per-repository means over \emph{all} frozen cells, including the two tasks that neither branch solved (4 tasks each). Restricted to verified-endpoint tasks the advantages are \DjangoAdvantageVer\,s, \SklearnAdvantageVer\,s, and \SympyAdvantageVer\,s respectively; the SymPy figure below is inflated by the unsolved \texttt{sympy-23950}. Shown directly rather than bootstrapped: three repositories cannot support a repository-level resampling interval.}
\label{tab:per-repo}
\begin{tabular}{lrrrrr}
\toprule
Repository & Tasks & CP (s) & SKIP (s) & Advantage (s) & Verified \\
\midrule
django/django & 4 & \DjangoCP & \DjangoSKIP & \DjangoAdvantage & 4/4 \\
scikit-learn/scikit-learn & 4 & \SklearnCP & \SklearnSKIP & \SklearnAdvantage & 3/4 \\
sympy/sympy & 4 & \SympyCP & \SympySKIP & \SympyAdvantage & 3/4 \\
\bottomrule
\end{tabular}
\end{table}

\subsection{Secondary cost measures}
Wall-clock advantage is the primary outcome, but the same interventions moved the token and tool-call measures in the same direction, which is a useful consistency check on the pairing. On verified-endpoint tasks, \textsc{CP} re-executed \VerTokenAdvantage{} fewer tokens and \VerToolCallAdvantage{} fewer tool calls per task on average than \textsc{Skip}. The gap tracks the decomposition closely: the first-checkpoint contrast saved \FirstTokenSaving{} tokens and 14.4 tool calls per task, the additional-checkpoint contrast only 17{,}530 tokens and 0.6 tool calls. We report these as separate observed quantities and do not combine them with wall-clock time into a single weighted cost.

Semantic rediscovery and divergent rework were not manually annotated in this pilot. The harness records fields for both, but they are zero throughout the real run rather than measured, and must not be read as evidence that no rediscovery occurred.

\section{Threats to validity and reproducibility}
\label{app:threats}
\paragraph{Counterfactual nondeterminism.} Model stochasticity can make paired recoveries diverge, and the agent runtime exposes no controllable sampling seed, so branch-order randomization and repetition, not seed matching, are what control it. Advantage is therefore an expected value with task-level uncertainty, and individual pair differences can change sign; \StabilizedCells{} of the \NumCandidatesN{} cells did, and each received a third repetition under the predeclared rule rather than being discarded.

\paragraph{Small effective sample size.} The \NumCandidatesN{} candidate cells are not \NumCandidatesN{} independent samples. Confidence intervals resample tasks: the \VerTasks{} verified-endpoint tasks for the first-checkpoint, additional-checkpoint, and pooled estimates, and all \NumTasksN{} for the termination-cost sensitivity. Repository results are shown directly. The workload is one difficulty tier of one benchmark, in one language, executed by one agent and one model version, and the reported effect sizes should not be extrapolated beyond that.

\paragraph{Single-shot failure sampling.} Each candidate carries one frozen logical failure ($K=1$ in Appendix~\ref{app:oracle}), so measured advantage is conditional on that failure location rather than averaged over a failure distribution. Fallback identity enters as a stratification across candidates instead. A study characterizing $\mathcal{F}$ would need multiple offsets per candidate.

\paragraph{Failure-model specificity.} Fail-stop failures isolate placement but omit semantic corruption and irreversible side effects; those require complementary recovery semantics.

\paragraph{Scope of claims.} This paper reports a measurement. It does not claim that \RippleCP{} outperforms any checkpoint-placement heuristic, that checkpoint advantage can be predicted on held-out tasks, or that ripple features are more informative than physical state size. Those experiments have not been run on this workload.

\paragraph{Reproducibility.} Released or archived with the paper: exact task IDs, base commits, and benchmark checksum; agent, model version, and sampling configuration (\PrimaryAgentModel); checkpoint backend and contents (\CheckpointBackendShort); candidate-boundary rules and the frozen stratification; failure locations; the paired-recovery harness and its preflight checks; branch-order and repetition seeds; the per-pair audit records and the list of cells that received a third repetition; the \BootstrapReps-replicate bootstrap procedure and its seed (\BootstrapSeed); and the analysis code that produces every table and figure in this paper. Total experiment cost was \TotalTokens{} re-executed tokens, \AgentHours{} agent-hours, and \BranchHours{} branch-hours of recovery execution (excluding checkpoint creation) across \NumBranches{} branches; monetary cost is not reported because the agent was run under an interactive subscription that does not expose per-run billing.

\section{Motivating intuition}
\label{app:intuition}
\begin{figure}[h]
\centering
\resizebox{\linewidth}{!}{%
\begin{tikzpicture}[
    font=\small,
    >=Latex,
    state/.style={
        draw,
        rounded corners=2pt,
        align=center,
        minimum height=7mm,
        text width=2.7cm,
        inner sep=3pt
    },
    event/.style={
        draw,
        rounded corners=2pt,
        align=center,
        minimum height=7mm,
        text width=1.8cm,
        inner sep=3pt
    },
    smallnode/.style={
        draw,
        rounded corners=2pt,
        align=center,
        minimum height=6mm,
        text width=1.25cm,
        inner sep=2pt,
        font=\scriptsize
    },
    decision/.style={
        draw,
        rounded corners=3pt,
        align=center,
        minimum height=8mm,
        text width=1.8cm,
        inner sep=3pt,
        font=\small\bfseries
    },
    arr/.style={
        ->,
        thick
    },
    dashedarr/.style={
        ->,
        thick,
        dashed
    }
]


\node[font=\bfseries] at (0,3.55)
{Large state change, low value};

\node[state] (download) at (0,2.30)
{\textbf{Large state change}\\
Download / build\\
1--2 GB};

\node[decision] (decisionL) at (0,0.75)
{Checkpoint\\now or skip?};

\node[event] (laterL) at (0,-0.65)
{Later\\execution};

\node[event] (failureL) at (0,-1.80)
{\textbf{Failure}};

\node[
    state,
    text width=2.9cm
] (redoL) at (0,-3.25)
{\textbf{Cheap recovery}\\
Re-download or rebuild\\
with little lost work};

\draw[arr] (download) -- (decisionL);

\draw[arr]
(decisionL) --
node[right,font=\scriptsize]{skip}
(laterL);

\draw[arr] (laterL) -- (failureL);
\draw[arr] (failureL) -- (redoL);


\node[smallnode] (l1) at (-4.15,2.30)
{Temporary\\artifact};

\node[smallnode] (l2) at (-4.15,1.05)
{Build\\cache};

\draw[dashedarr]
(download.west) -- ++(-0.30,0)
|- (l1.east);

\draw[dashedarr]
(download.west) -- ++(-0.25,0)
|- (l2.east);

\node[
    align=center,
    text width=2.6cm,
    font=\scriptsize
] at (-2.15,-0.10)
{\textbf{Small ripple}\\
Few future decisions\\
depend on this state};


\draw[densely dashed]
(3.0,3.80) -- (3.0,-4.45);


\node[font=\bfseries] at (6.9,3.55)
{Tiny state change, high value};

\node[
    state,
    text width=3.55cm
] (discovery) at (6.9,2.25)
{\textbf{Tiny semantic decision}\\
Library X is incompatible\\
with CUDA 12.6};

\node[decision] (decisionR) at (6.9,0.65)
{Checkpoint\\now or skip?};

\draw[arr] (discovery) -- (decisionR);


\node[smallnode] (dep) at (4.55,-0.85)
{Change\\dependency};

\node[smallnode] (config) at (6.10,-0.85)
{Update\\config};

\node[smallnode] (build) at (7.70,-0.85)
{Successful\\build};

\node[smallnode] (test) at (9.25,-0.85)
{Tests\\pass};


\coordinate (fan) at (6.9,-0.10);

\draw[thick]
(decisionR.south) -- (fan);

\draw[arr]
(fan) -| (dep.north);

\draw[arr]
(fan) -| (config.north);

\draw[arr]
(fan) -| (build.north);

\draw[arr]
(fan) -| (test.north);

%

\draw[dashedarr]
(dep.east) -- (config.west);

\draw[dashedarr]
(config.east) -- (build.west);

\draw[dashedarr]
(build.east) -- (test.west);


\node[event] (failureR) at (6.9,-2.00)
{\textbf{Later}\\
\textbf{failure}};

\node[
    state,
    text width=4.85cm
] (redoR) at (6.9,-3.45)
{\textbf{Expensive recovery if skipped}\\
Rediscover cause and repeat\\
reasoning, tools, edits, and tests};

\draw[arr]
(test.south) -- ++(0,-0.35)
|- (failureR.east);

\draw[arr]
(failureR.south) -- (redoR.north);


\node[
    align=center,
    text width=2.25cm,
    font=\scriptsize
] at (10.55,-2.00)
{\textbf{Large ripple}\\
Many future actions\\
depend on one decision};


\node[
    draw,
    rounded corners=3pt,
    align=center,
    text width=4.8cm,
    inner sep=5pt,
    font=\small
] at (3.0,-5.00)
{
\textbf{RippleCP:}
checkpoint based on future work protected,
not simply bytes changed.
};

\end{tikzpicture}%
}
\caption{
\textbf{RippleCP intuition.}
A large physical state change may have low checkpoint value when it is
cheap to reproduce (left), whereas a small semantic decision can have
high value when many subsequent actions depend on it (right).
RippleCP estimates whether checkpointing the current state avoids enough
future recovery cost to justify materializing it.
}
\label{fig:ripplecp-overview}
\end{figure}
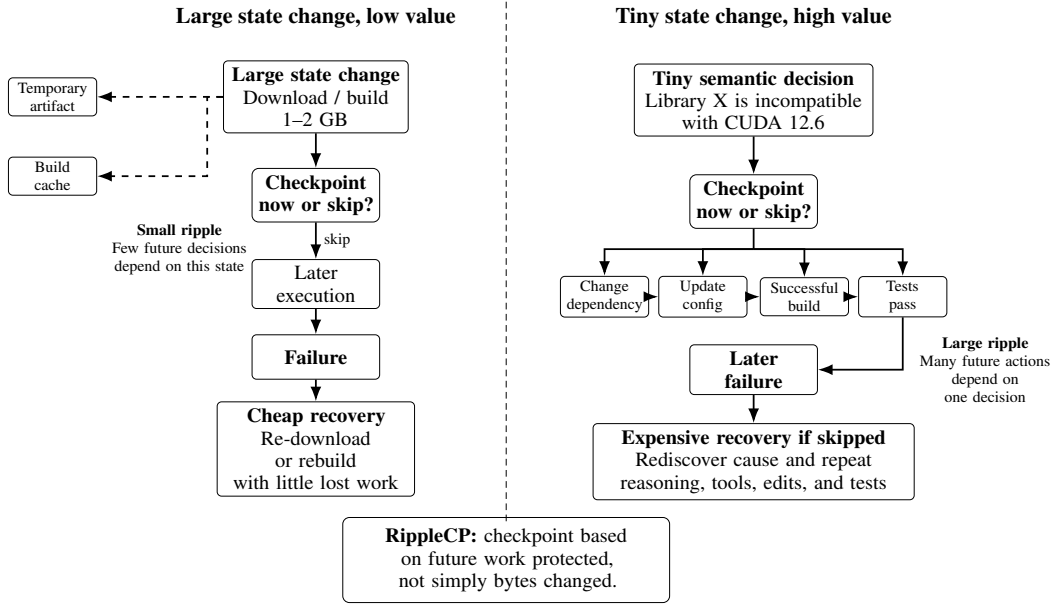

The placement problem this paper measures is motivated by a mismatch between how much state an
action changes and how much future work that state protects. Figure~\ref{fig:ripplecp-overview}
contrasts the two cases. The pilot in Section~\ref{sec:pilot} does not test this mechanism
directly, since it measures the advantage itself rather than its cause, so the figure is
included as motivation rather than as a result.

\end{document}